\documentclass{article} % For LaTeX2e
\usepackage{amsmath,amsfonts,bm}

\def\eqref#1{equation~\ref{#1}}
\def\1{\bm{1}}

\DeclareMathAlphabet{\mathsfit}{\encodingdefault}{\sfdefault}{m}{sl}
\SetMathAlphabet{\mathsfit}{bold}{\encodingdefault}{\sfdefault}{bx}{n}

\usepackage{hyperref}
\usepackage{url}
\usepackage{comment}
\usepackage{graphicx}
\usepackage{amsmath}
\newcommand{\norm}[1]{\left\lVert#1\right\rVert}

\title{CWAE-IRL: Formulating a supervised approach to Inverse Reinforcement Learning problem}

\author{Arpan Kusari \\
Department of Robotics and AI\\
Ford Motor Company\\
Dearborn, MI\\
\texttt{akusari@ford.com} \\
}

\begin{document}

\maketitle

\begin{abstract}
Inverse reinforcement learning (IRL) is used to infer the reward function from the actions of an expert running a Markov Decision Process (MDP). A novel approach using variational inference for learning the reward function is proposed in this research. Using this technique, the intractable posterior distribution of the continuous latent variable (the reward function in this case) is analytically approximated to appear to be as close to the prior belief while trying to reconstruct the future state conditioned on the current state and action. The reward function is derived using a well-known deep generative model known as Conditional Variational Auto-encoder (CVAE) with Wasserstein loss function, thus referred to as Conditional Wasserstein Auto-encoder-IRL (CWAE-IRL), which can be analyzed as a combination of the backward and forward inference. This can then form an efficient alternative to the previous approaches to IRL while having no knowledge of the system dynamics of the agent. Experimental results on standard benchmarks such as objectworld and pendulum show that the proposed algorithm can effectively learn the latent reward function in complex, high-dimensional environments.
\end{abstract}

\section{Introduction}

%Origin
Reinforcement learning, formalized as Markov decision process (MDP), provides a general solution to sequential decision making, where given a state, the agent takes an optimal action by maximizing the long-term reward from the environment \cite{bellman1957markovian}. However, in practice, defining a reward function that weighs the features of the state correctly can be challenging, and techniques like reward shaping are often used to solve complex real-world problems \cite{ng1999policy}. The process of inferring the reward function given the demonstrations by an expert is defined as inverse reinforcement learning (IRL) or apprenticeship learning  \cite{ng2000algorithms,abbeel2004apprenticeship}.

% Problems of IRL
The fundamental problem with IRL lies in the fact that the algorithm is under defined and infinitely different reward functions can yield the same policy \cite{finn2016guided}. Previous approaches have used preferences on the reward function to address the non-uniqueness. \cite{ng2000algorithms} suggested reward function that maximizes the difference in the values of the expert's policy and the second best policy. \cite{ziebart2008maximum} adopted the principle of maximum entropy for learning the policy whose feature expectations are constrained to match those of the expert's. \cite{ratliff2006maximum} applied the structured max-margin optimization to IRL and proposed a method for finding the reward function that maximizes the margin between expert's policy and all other policies. \cite{neu2009training} unified a direct method that minimizes deviation from the expert's behavior and an indirect method that finds an optimal policy from the learned reward function using IRL. \cite{syed2008game} used a game-theoretic framework to find a policy that improves with respect to an expert's. 

Another challenge for IRL is that some variant of the forward reinforcement learning problem needs to be solved in a tightly coupled manner to obtain the corresponding policy, and then compare this policy to the demonstrated actions \cite{finn2016guided}.  Most early IRL algorithms proposed solving an MDP in the inner loop \cite{ng2000algorithms,abbeel2004apprenticeship,ziebart2008maximum}. This requires perfect knowledge of the  expert's dynamics which are almost always impossible to have. Several works have proposed to relax this requirement, for example by learning a value function instead of a cost \cite{todorov2007linearly}, solving an approximate local control problem \cite{levine2012continuous} or generating a discrete graph of states \cite{byravan2015graph}. However, all these methods still require some partial knowledge of the system dynamics.

% IRL implementations
% Linearity vs non-linearity of reward functions
Most of the early research in this field has expressed the reward function as a weighted linear combination of hand selected features \cite{ng2000algorithms, ramachandran2007bayesian, ziebart2008maximum}. Non-parametric methods such as Gaussian Processes (GPs) have also been used for potentially complex, nonlinear reward functions \cite{levine2011nonlinear}. While in principle this helps extend the IRL paradigm to flexibly account for non-linear reward approximation; the use of kernels simultaneously leads to higher sample size requirements. Universal function approximators such as non-linear deep neural network have been proposed recently \cite{wulfmeier2015deep,finn2016guided}. This moves away from using hand-crafted features and helps in learning highly non-linear reward functions but they still need the agent in the loop to generate new samples to "guide" the cost to the optimal reward function. \cite{fu2017learning} has recently proposed deriving an adversarial reward learning formulation which disentangles the reward learning process by a discriminator trained via binary regression data and uses policy gradient algorithms to learn the policy as well. 

%Motivation of current approach
%How can bayesian IRL help? What has been lacking in other approaches? 
The Bayesian IRL (BIRL) algorithm proposed by \cite{ramachandran2007bayesian} uses the expert's actions as evidence to update the prior on reward functions. The reward learning and apprenticeship learning steps are solved by performing the inference using a modified Markov Chain Monte Carlo (MCMC) algorithm. \cite{zheng2014robust} described an expectation-maximization (EM) approach for solving the BIRL problem, referring to it as the Robust BIRL (RBIRL). Variational Inference (VI) has been used as an efficient and alternative strategy to MCMC sampling for approximating posterior densities \cite{jordan1999introduction,wainwright2008graphical}. Variational Auto-encoder (VAE) was proposed by \cite{kingma2014stochastic} as a neural network version of the approximate inference model. The loss function of the VAE is given in such a way that it automatically tries to maximize the likelihood of the data given the current latent variables (reconstruction loss), while encouraging the latent variables to be close to our prior belief of how the variables should look like (Kullbeck-Liebler divergence loss). This can be seen as an generalization of EM from maximum a-posteriori (MAP) estimation of the single parameter to an approximation of complete posterior distribution. Conditional VAE (CVAE) has been proposed by \cite{sohn2015learning} to develop a deep conditional generative model for structured output prediction using Gaussian latent variables. Wasserstein Auto-Encoder (WAE) has been proposed by \cite{tolstikhin2017wasserstein} to utilize Wasserstein loss function in place of KL divergence loss for robustly estimating the loss in case of small samples, where VAE fails. 

% Novel contributions
This research is motivated by the observation that IRL can be formulated as a supervised learning problem with latent variable modelling. This intuition is not unique. It has been proposed by \cite{klein2013cascaded} using the Cascaded Supervised IRL (CSI) approach. However, CSI uses non-generalizable heuristics to classify the dataset and find the decision rule to estimate the reward function. Here, I propose to utilize the CVAE framework with Wasserstein loss function to determine the non-linear, continuous reward function utilizing the expert trajectories without the need for system dynamics. The encoder step of the CVAE is used to learn the original reward function from the next state conditioned on the current state and action. The decoder step is used to recover the next state given the current state, action and the latent reward function. 

The likelihood loss, composed of the reconstruction error and the Wasserstein loss, is then fed to optimize the CVAE network. The Gaussian distribution is used here as the prior distribution; however, \cite{ramachandran2007bayesian} has described various other prior distributions which can be used based on the class of problem being solved. Since, the states chosen are supplied by the expert's trajectories, the CWAE-IRL algorithm is run only on those states without the need to run an MDP or have the agent in the loop. 
Two novel contributions are made in this paper:
\begin{itemize}
\item Proposing a generative model such as an auto-encoder for estimating the reward function leads to  a more effective and efficient algorithm with locally optimal, analytically approximate solution.
\item Using only the expert's state-action trajectories provides a robust generative solution without any knowledge of system dynamics. 
\end{itemize}
Section \ref{sec:prelim} gives the background on the concepts used to build our model; Section \ref{sec:method} describes the proposed methodology; Section \ref{sec:results} gives the results and Section \ref{sec:conclusions} provides the discussion and conclusions.  

\section{Preliminaries}
\label{sec:prelim}

\subsection{Markov Decision Process (MDPs)}
In the reinforcement learning problem, at time t, the agent observes a state, $s_t \in$ S, and takes an action, $a_t \in$ A; thereby receiving an immediate scalar reward $r_t$ and moving to a new state $s_{t+1}$. The model's dynamics are characterized by state transition probabilities $p(s_{t+1}|s_t,a_t)$. This can be formally stated as a Markov Decision Process (MDP) where the next state can be completely defined by the previous state and action (Markov property) and the agent receives a scalar reward for executing the action \cite{bellman1957markovian}.

The goal of the agent is to maximize the cumulative reward (discounted sum of rewards) or value function:

\begin{equation}
\label{eq:value_func}
v_t = \sum_{k=0}^{\infty} \gamma^k r_{t+k}
\end{equation} 

where $0 \leq \gamma \leq 1$ is the discount factor and $r_t$ is the reward at time-step $t$. 

In terms of a policy $\pi : S \rightarrow A$, the value function can be given by Bellman equation as:

\begin{align}
\label{eq:bellman}
v_{\pi}(s_t) &= \mathop{\mathbb{E}}_{\pi} \Bigg[  \sum_{k=0}^{\infty} \gamma^k r_{t+k} | S = s_t \Bigg] \\
 &= \mathop{\mathbb{E}}_{\pi} \Bigg[ r_t + \sum_{k=1}^{\infty} \gamma^k r_{t+k} | S = s_t \Bigg] \\
 &\begin{aligned}
 &= \sum_{a} \pi (a_t|s_t) \sum_{s_{t+1}} p(s_{t+1}|s_t,a_t)\\
 &\qquad \Bigg[ r(s_t,a_t,s_{t+1}) + \gamma \mathop{\mathbb{E}}_{\pi}  \Bigg[\sum_{k=0}^{\infty} \gamma^k r_{t+k} | S=s_{t+1} \Bigg] \Bigg] 
 \end{aligned} \\
 &\begin{aligned}
 \label{eq:bellman2}
 &= \sum_{a} \pi (a_t|s_t) \sum_{s_{t+1}} p(s_{t+1}|s_t,a_t)\\
 &\qquad \Bigg[ r(s_t,a_t,s_{t+1}) + \gamma v_{\pi}(s_t+1) \Bigg] 
 \end{aligned}
\end{align}
Using Bellman's optimality equation, we can define, for any MDP, a policy $\pi$ is greater than or equal to any other policy $\pi'$ if value function $v_{\pi}(s_t) \leq v_{\pi'}(s_t)$ for all $s_t \in$ S. This policy is known as an optimal policy ($\pi^*$) and its value function is known as optimal value function ($v^*$).

\subsection{Bayesian IRL}
\label{subsec:birl}
The bayesian approach to IRL was proposed by \cite{ramachandran2007bayesian} by encoding the reward function preference as a prior and optimal confidence of the behavior data as the likelihood. Considering the expert $\chi$ is executing an MDP $M = (S,A,p,\gamma)$, the reward for $\chi$ is assumed to be sampled from a prior (known) distribution $P_R$ defined as:
\begin{equation}
P_R(R) = \prod_{s \in S, a \in A} P(R(s,a))
\end{equation}
The distribution to be used as a prior depends on the type of problem. The expert's goal of maximizing accumulated reward is equivalent to finding the optimal action of each state. The likelihood thus defines our confidence in $\chi$'s ability to select the optimal action. This is modeled as a exponential  distribution for the likelihood of trajectory $\tau$ with $Q^*$ as:
\begin{equation}
Pr_{\chi}(\tau|R) = \dfrac{1}{Z'} \exp^{\alpha_{\chi}Q^* (\tau,R)}
\end{equation}
where $\alpha_{\chi}$ is a parameter representing the degree of confidence in $\chi$'s ability. 
The posterior probability of the reward function $R$ is computed using Bayes theorem, 
\begin{align}
Pr_{\chi}(R|\tau) &= \dfrac{Pr_{\chi}(\tau|R)P_R(R)}{Pr(\tau)}\\
                  &= \dfrac{1}{Z}\exp^{\alpha_{\chi}Q^* (\tau,R)}P_R(R)
\end{align}
BIRL uses MCMC sampling to compute the posterior mean of the reward function. 

\subsection{Variational Inference}
\label{subsec:vi}
For observations $x=x_{1:n}$ and latent variables $z=z_{1:m}$, the joint density can be written as:
\begin{equation}
p(z,x) = p(z)p(x|z)
\end{equation}
The latent variables are drawn from a prior distribution $p(z)$ and they are then related to the observations through the likelihood $p(x|z)$. Inference in a bayesian framework amounts to conditioning on data and computing the posterior $p(z|x)$. In lot of cases, this posterior is intractable and requires approximate inference. Variational inference has been proposed in the recent years as an alternative to MCMC sampling by using optimization instead of sampling \cite{blei2017variational}.  

For a family of approximate densities $\zeta$ over the latent variables, we try to find a member of the family that minimizes the Kullback-Leibler (KL) divergence to the exact posterior
\begin{equation}
q^*(z) = \underset{q(z) \in \zeta}{arg min} KL(q(z)||p(z|x))
\end{equation}
The posterior is then approximated with the optimized member of the family $q^*(z)$. The KL divergence is then given by
\begin{align}
KL(q(z)||p(z|x)) &= \mathbb{E}[log(q(z))]-\mathbb{E}[log(p(z|x))] \\
&= \mathbb{E}[log(q(z))]-\mathbb{E}[log(p(z,x))]+log(p(x))
\end{align}
Since, the divergence cannot be computed, an alternate objective is optimized in VAE called evidence lower bound (ELBO) that is equivalent,
\begin{align}
L_{ELBO}(q) &= \mathbb{E}[log(p(x))] - KL(q(z)||p(z|x)\\
&= \mathbb{E}[log(p(x|z))] - KL(q(z|x)||p(z|x)
\end{align}
This can be defined as a sum of two separate losses:
\begin{equation}
\label{eq:vi}
L_{VAE} = L_{lk} + L_{div}
\end{equation}
where $L_{lk}$ is the loss related to the log-likelihood and $L_{div}$ is the loss related to the divergence.

CVAE is used to perform probabilistic inference and predict diversely for structured outputs. The loss function is slightly altered with the introduction of class labels $c$:
\begin{align}
    \label{eq:cvae_loss}
    L_{CVAE} &= \mathbb{E}[log(p(x|c)] - KL(q(z|x,c)||p(z|x,c))\\
    &= \mathbb{E}[log(p(x|z,c))] - KL(q(z|x,c)||p(z|x,c))
\end{align}

\subsection{Wasserstein loss function}
Wasserstein distance, also known as Kantorovich-Rubenstein distance or earth mover's distance (EMD) \cite{rubner2000earth}, provides a natural distance over probability distributions in the metric space \cite{frogner2015learning}. It is a formulation of optimal transport problem \cite{villani2008optimal} where the Wasserstein distance is the minimum cost required to move a pile of earth (an arbitrary distribution) to another. The mathematical formulation given by Kantorovich \cite{tolstikhin2017wasserstein} is:
\begin{equation}
    W_c(P_X, P_Y) = \underset{\Gamma \in P(X \sim P_X, Y \sim P_Y)}{inf} \mathbb{E}_{(X, Y)\sim \Gamma} [c(X,Y)]
\end{equation}
where $c(X,Y)$ is the cost function, $X$ and $Y$ are random variables with marginal distributions $P_X$ and $P_Y$ respectively. 

EMD has been utilized in various practical applications in computer science such as pattern recognition in images \cite{he2018wasserstein}. Wasserstein GAN (WGAN) has been proposed by \cite{arjovsky2017wasserstein} to minimize the EMD between the generative distribution and the data distribution. \cite{tolstikhin2017wasserstein} proposed Wasserstein Auto-encoder (WAE) where the divergence loss has been calculated using the EMD instead of KL-divergence and has been shown to be robust in presence of noise and smaller samples.

\section{Conditional Wasserstein Auto-Encoder-IRL}
\label{sec:method}

\begin{figure}[h!]
    \centering
        \includegraphics[width=\columnwidth]{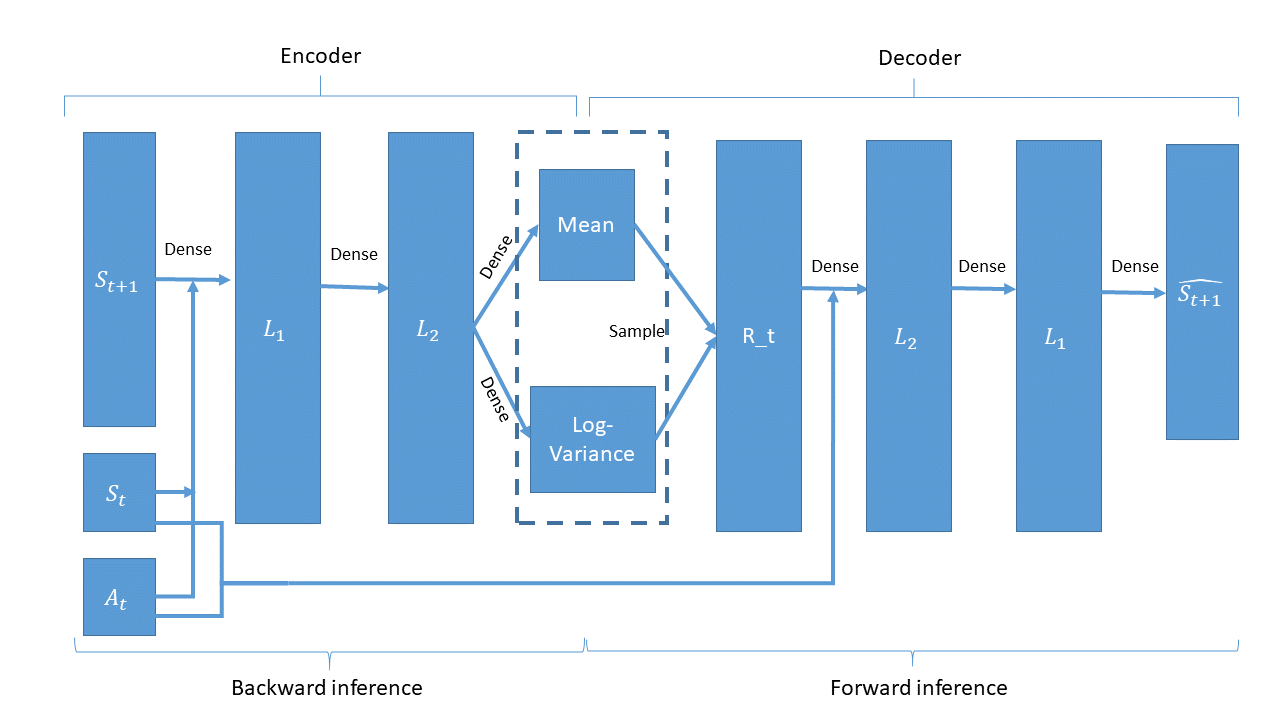}
        \caption{CWAE-IRL Network architecture}
        \label{fig:cvae_irl}
\end{figure}

In this paper, my primary argument is that the inverse reinforcement learning problem can be devised as a supervised learning problem with learning of latent variable. The reward function, $r(s_t, a_t, s_{t+1})$, can be formulated as a latent function which is dependent on the state at time $t$, $s_t$, action at time $t$, $a_t$, and state at time $(t+1)$, $s_{t+1}$. In the CVAE framework, using the state and action pair as the class label $c$ and rewriting the CVAE loss in Equation \ref{eq:cvae_loss} with $s_{t+1}$ as $x$ and reward at time $t$, $r_t$ as $z$, we get:
\begin{equation}
    \label{eq:cvae_irl_loss}
    L_{CVAE-IRL} = \mathbb{E}[log(p(s_{t+1}|s_t, a_t))] - KL(q(r_t|s_{t+1},s_t, a_t)||p(r_t|s_{t+1},s_t, a_t))
\end{equation}
The first part of Equation \ref{eq:cvae_irl_loss} provides the log likelihood of transition probability of an MDP and the second part gives the KL-divergence of the encoded reward function to the prior gaussian belief. Thus, the proposed method tries to recover the next state from the current state and current action by encoding the reward function as the latent variable and constraining the reward function to lie as close to the gaussian function. The network structure of the method is given in Figure \ref{fig:cvae_irl}. 

\subsection{Encoder}
The encoder is a neural network which inputs the current state, current action and next state and generates a probability distribution $q(r_t|s_{t+1}, s_t, a_t)$, assumed to be isotropic gaussian. Since a near-optimal policy is inputted into the IRL framework, minibatches of randomly sampled $(s_{t+1}, s_t, a_t)$ are introduced into the network. Two hidden layers with dropout are used to encode the input data into a latent space, giving two outputs corresponding to the mean and log-variance of the distribution. This step is similar to the backward inference problem in IRL methodology where the reward function is constructed from the sampled trajectories for a near-optimal agent. 

\subsection{Decoder}
Given the current state and action, the decoder maps the latent space into the state space and reconstructs the next state $\hat{s}_{t+1}$ from a sampled $r_t$ (from the normal distribution). Similar to the VAE formulation, samples are generated from a standard normal distribution and reparameterized using the mean and log-variance computed in the encoder step. This step resembles the forward inference problem of an MDP where given a state, action and reward distribution, we estimate the next state that the agent gets to. Two hidden layers with dropout are used similar to the encoder. 

\subsection{Using Wasserstein loss}
Even though the KL-divergence should be able to provide for the loss theoretically, it does not converge in practice and indeed gives really large values in case of small samples such as in our formulation. \cite{tolstikhin2017wasserstein} provides a Maximum Mean Discrepancy (MMD) measure based on Wasserstein metric for a positive-definite reproducing kernel $k(\cdot,\cdot)$ such as the Radial Basis Function (RBF) kernel:
\begin{equation}
    MMD_k(P_z, Q_z) = \norm{\int_z k(z, \cdot)dP_z(z) - \int_z k(z, \cdot)dQ_z(z)}_{H_k}
\end{equation}
where $H_k$ is the Reproducing Kernel Hilbert Space (RKHS) mapping $z : \mathbb{Z} \rightarrow \mathbb{R}$. The divergence loss can then be written as:
\begin{multline}
 W(q(z|x)||p(z|x)) = \frac{\lambda}{n(n-1)}\sum_{l \neq j}k(z_l, z_j) + \frac{\lambda}{n(n-1)}\sum_{l \neq j}k(\Tilde{z_l}, \Tilde{z_j}) \\ -\frac{2\lambda}{n^2}\sum_{l,j}k(z_l, \Tilde{z_j})
\end{multline}
where $c$ is the cost between the input, $x$ and the output,  $\Tilde{z}$, of the decoder, $D$, using the sampled latent variable, $z$ (given as mean squared error), The resulting CWAE-IRL loss function is given as:
\begin{equation}
    L_{CWAE-IRL} = \mathbb{E}[log(p(s_{t+1}|s_t,a_t))] - W(q(r_t|s_{t+1},s_t,a_t)||p(r_t|s_{t+1}, s_t, a_t))
\end{equation}

\section{Experiments}
\label{sec:results}
In this section, I present the results of CWAE-IRL on two simulated tasks, objectworld and pendulum. 
\subsection{Objectworld}
Objectworld is a generalization of gridworld \cite{sutton1998reinforcement}, described in \cite{levine2011nonlinear}. 
It contains NxN grid of states with five actions per state, corresponding to steps in each direction and staying in place. Each action has a 30\% chance of moving in a different random direction. There are randomly assigned objects, having one of 2 inner and outer colors chosen, red and green. There are 4 continuous features for each of the grids, each giving the Euclidean distance to the nearest object with a specific inner or outer color. The true reward is +1 in states within 3 cells of outer red and 2 cells of outer green, -1 within 3 cells of outer red, and zero otherwise. Inner colors serve as distractors. The expert trajectories fed have a sample length of 16. The algorithms for objectworld, Maximum Entropy IRL and Deep Maximum Entropy IRL are used from the GitHub implementation of \cite{alger2016irl} without any modifications. Only continuous features are used for all implementations. 

\begin{figure}[h!]
   \includegraphics[width=\columnwidth]{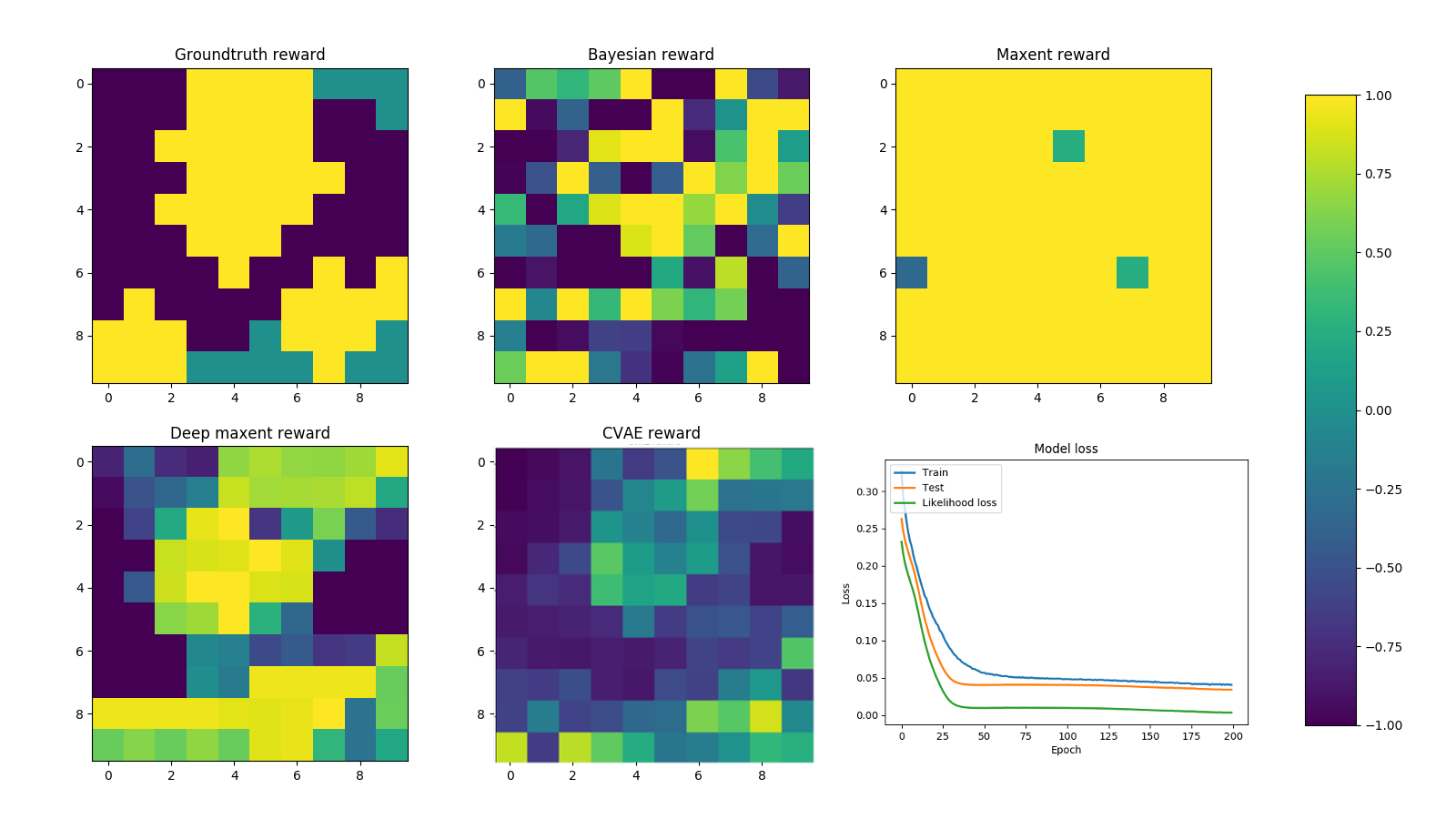}
   \caption{Reward comparison of IRL algorithms for an objectworld of grid size 10 with random placement of objects. (a) Groundtruth reward (b) Bayesian IRL reward (c) Maximum entropy reward (d) Deep maximum entropy reward (e) VAE reward (f) Training loss (blue), validation loss (red) and likelihood loss (green) over epochs}
  \centering
  \label{fig:reward}
\end{figure}

CWAE-IRL is compared with prior IRL methods, described in the previous sections. Among prior methods chosen, CWAE-IRL is compared to BIRL \cite{ramachandran2007bayesian}, Maximum Entropy IRL \cite{ziebart2008maximum} and Deep Maximum Entropy IRL \cite{wulfmeier2015deep}. Only the Maximum Entropy IRL uses the reward as a linear combination of features while the others describe it in a non-linear fashion. The learnt rewards for all the algorithms are shown in Figure \ref{fig:reward} with an objectworld of grid size 10. CWAE-IRL can recover the original reward distribution while the Deep Maximum Entropy IRL overestimates the reward in various places.  Maximum Entropy IRL and BIRL completely fail to learn the rewards. Deep Maximum Entropy tends to give negative rewards to state spaces which are not well traversed in the example trajectories. However, CWAE-IRL generalizes over the spaces even though they have not been visited frequently. Also, due to the constraint of being close to the prior gaussian belief, the scale of rewards are best captured by the proposed algorithm as compared to the other algorithms which tend to overscale the rewards. 

\subsection{Pendulum}
The pendulum environment \cite{brockman2016openai} is an well-known problem in the control literature in which a pendulum starts from a random position and the goal is to keep it upright while applying the minimum amount of force. The state vector is composed of the cosine (and sine) of the angle of the pendulum, and the derivative of the angle. The action is the joint effort as $11$ discrete actions linearly spaced within the $[-2, 2]$ range. The reward is 
\begin{equation}
    \label{eq:pendulum_reward}
    R = -(w_1 \cdot \norm{\theta}^2 + w_2 \cdot \norm{\dot{\theta}}^2 + w_3 \cdot \norm{a}^2),
\end{equation}
where $w_1$, $w_2$ and $w_3$ are the reward weights for the angle $\theta$, derivative of angle $\dot{\theta}$ and action $a$ respectively. The optimal reward weights given by OpenAI are $[1, 0.1, 0.001]$ respectively. An episode is limited to 1000 timesteps. 

\begin{figure}[h!]
   \includegraphics[width=\columnwidth]{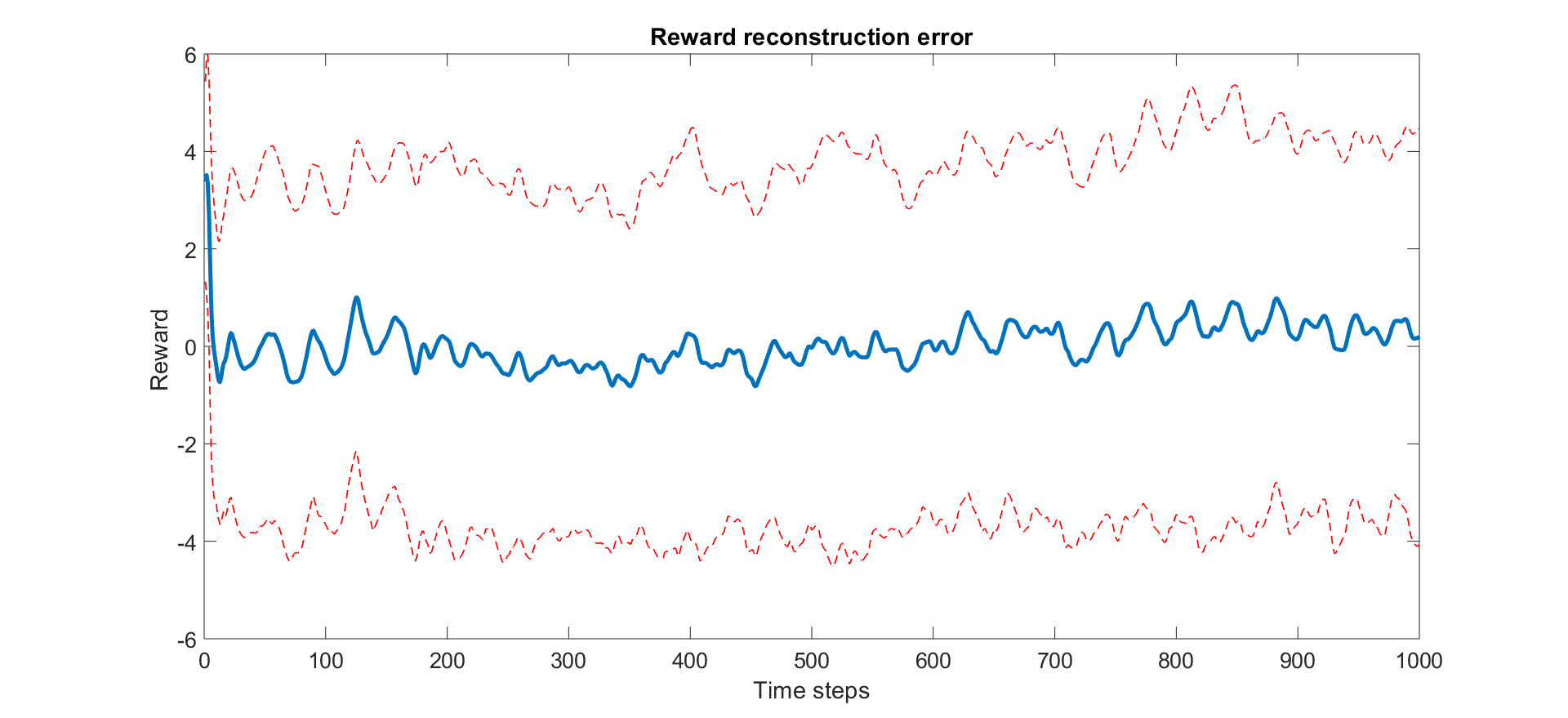}
   \caption{Reward error for predicted minus the actual error recovered at each time step (smoothed) with 1 $\sigma$ confidence interval for the pendulum environment}
  \centering
  \label{fig:pendulum_reward}
\end{figure}

% Solving with DQN
A deep Q-network (DQN) has been proposed by \cite{mnih2015human} that combines deep neural networks with RL to solve continuous state discrete action problems. DQN uses a neural network with gives the Q-values for every action and uses a buffer to store old states and actions to sample from to help stabilize training. Using a continuous state space makes it impossible to have all states visited during training. This also makes it very difficult for the comparison of recovered reward with the actual reward. The DQN is trained for 50,000 episodes. The CWAE-IRL is trained using 25 trajectories and the reward is predicted for 5 trajectories. The error plot between the reward recovered and the actual reward is given in Figure \ref{fig:pendulum_reward}. The mean error hovers around 0 showing that under for majority of the states and actions, the proposed method is able to recover the correct reward. 

\section{Conclusions}
\label{sec:conclusions}
I have presented an algorithm for inverse reinforcement learning which learns the latent reward function using a conditional variational auto-encoder with Wasserstein loss function. It learns the reward function as a continuous, Gaussian distribution while trying to reconstruct the next state given the current state and action. The proposed model makes the inference process scalable while making it easier for inference of reward functions.  Inferring a continuous parametric distribution of the reward functions can prove useful for classifying behaviors of decision making agents in diverse applications such as autonomous driving. 

\subsubsection*{Acknowledgments}

I would like to acknowledge the help of Satabdi Saha in editing the manuscript. 

\bibliography{iclr2020_conference}

\begin{thebibliography}{10}
\providecommand{\url}[1]{#1}
\csname url@rmstyle\endcsname
\providecommand{\newblock}{\relax}
\providecommand{\bibinfo}[2]{#2}
\providecommand\BIBentrySTDinterwordspacing{\spaceskip=0pt\relax}
\providecommand\BIBentryALTinterwordstretchfactor{4}
\providecommand\BIBentryALTinterwordspacing{\spaceskip=\fontdimen2\font plus
\BIBentryALTinterwordstretchfactor\fontdimen3\font minus
  \fontdimen4\font\relax}
\providecommand\BIBforeignlanguage[2]{{%
\expandafter\ifx\csname l@#1\endcsname\relax
\typeout{** WARNING: IEEEtran.bst: No hyphenation pattern has been}%
\typeout{** loaded for the language `#1'. Using the pattern for}%
\typeout{** the default language instead.}%
\else
\language=\csname l@#1\endcsname
\fi
#2}}

\bibitem{bellman1957markovian}
R.~Bellman, ``A markovian decision process,'' \emph{Journal of Mathematics and
  Mechanics}, pp. 679--684, 1957.

\bibitem{ng1999policy}
A.~Y. Ng, D.~Harada, and S.~Russell, ``Policy invariance under reward
  transformations: Theory and application to reward shaping,'' in \emph{ICML},
  vol.~99, 1999, pp. 278--287.

\bibitem{ng2000algorithms}
A.~Y. Ng, S.~J. Russell, \emph{et~al.}, ``Algorithms for inverse reinforcement
  learning.'' in \emph{Icml}, 2000, pp. 663--670.

\bibitem{abbeel2004apprenticeship}
P.~Abbeel and A.~Y. Ng, ``Apprenticeship learning via inverse reinforcement
  learning,'' in \emph{Proceedings of the twenty-first international conference
  on Machine learning}.\hskip 1em plus 0.5em minus 0.4em\relax ACM, 2004, p.~1.

\bibitem{finn2016guided}
C.~Finn, S.~Levine, and P.~Abbeel, ``Guided cost learning: Deep inverse optimal
  control via policy optimization,'' in \emph{International Conference on
  Machine Learning}, 2016, pp. 49--58.

\bibitem{ziebart2008maximum}
B.~D. Ziebart, A.~L. Maas, J.~A. Bagnell, and A.~K. Dey, ``Maximum entropy
  inverse reinforcement learning.'' in \emph{AAAI}, vol.~8.\hskip 1em plus
  0.5em minus 0.4em\relax Chicago, IL, USA, 2008, pp. 1433--1438.

\bibitem{ratliff2006maximum}
N.~D. Ratliff, J.~A. Bagnell, and M.~A. Zinkevich, ``Maximum margin planning,''
  in \emph{Proceedings of the 23rd international conference on Machine
  learning}.\hskip 1em plus 0.5em minus 0.4em\relax ACM, 2006, pp. 729--736.

\bibitem{neu2009training}
G.~Neu and C.~Szepesv{\'a}ri, ``Training parsers by inverse reinforcement
  learning,'' \emph{Machine learning}, vol.~77, no. 2-3, p. 303, 2009.

\bibitem{syed2008game}
U.~Syed and R.~E. Schapire, ``A game-theoretic approach to apprenticeship
  learning,'' in \emph{Advances in neural information processing systems},
  2008, pp. 1449--1456.

\bibitem{todorov2007linearly}
E.~Todorov, ``Linearly-solvable markov decision problems,'' in \emph{Advances
  in neural information processing systems}, 2007, pp. 1369--1376.

\bibitem{levine2012continuous}
S.~Levine and V.~Koltun, ``Continuous inverse optimal control with locally
  optimal examples,'' \emph{arXiv preprint arXiv:1206.4617}, 2012.

\bibitem{byravan2015graph}
A.~Byravan, M.~Monfort, B.~D. Ziebart, B.~Boots, and D.~Fox, ``Graph-based
  inverse optimal control for robot manipulation.'' in \emph{Ijcai}, vol.~15,
  2015, pp. 1874--1890.

\bibitem{ramachandran2007bayesian}
D.~Ramachandran and E.~Amir, ``Bayesian inverse reinforcement learning,''
  \emph{Urbana}, vol.~51, no. 61801, pp. 1--4, 2007.

\bibitem{levine2011nonlinear}
S.~Levine, Z.~Popovic, and V.~Koltun, ``Nonlinear inverse reinforcement
  learning with gaussian processes,'' in \emph{Advances in Neural Information
  Processing Systems}, 2011, pp. 19--27.

\bibitem{wulfmeier2015deep}
M.~Wulfmeier, P.~Ondruska, and I.~Posner, ``Deep inverse reinforcement
  learning,'' \emph{CoRR, abs/1507.04888}, 2015.

\bibitem{fu2017learning}
J.~Fu, K.~Luo, and S.~Levine, ``Learning robust rewards with adversarial
  inverse reinforcement learning,'' \emph{arXiv preprint arXiv:1710.11248},
  2017.

\bibitem{zheng2014robust}
J.~Zheng, S.~Liu, and L.~M. Ni, ``Robust bayesian inverse reinforcement
  learning with sparse behavior noise.'' in \emph{AAAI}, 2014, pp. 2198--2205.

\bibitem{jordan1999introduction}
M.~I. Jordan, Z.~Ghahramani, T.~S. Jaakkola, and L.~K. Saul, ``An introduction
  to variational methods for graphical models,'' \emph{Machine learning},
  vol.~37, no.~2, pp. 183--233, 1999.

\bibitem{wainwright2008graphical}
M.~J. Wainwright, M.~I. Jordan, \emph{et~al.}, ``Graphical models, exponential
  families, and variational inference,'' \emph{Foundations and
  Trends{\textregistered} in Machine Learning}, vol.~1, no. 1--2, pp. 1--305,
  2008.

\bibitem{kingma2014stochastic}
D.~P. Kingma and M.~Welling, ``Stochastic gradient vb and the variational
  auto-encoder,'' in \emph{Second International Conference on Learning
  Representations, ICLR}, 2014.

\bibitem{sohn2015learning}
K.~Sohn, H.~Lee, and X.~Yan, ``Learning structured output representation using
  deep conditional generative models,'' in \emph{Advances in neural information
  processing systems}, 2015, pp. 3483--3491.

\bibitem{tolstikhin2017wasserstein}
I.~Tolstikhin, O.~Bousquet, S.~Gelly, and B.~Schoelkopf, ``Wasserstein
  auto-encoders,'' \emph{arXiv preprint arXiv:1711.01558}, 2017.

\bibitem{klein2013cascaded}
E.~Klein, B.~Piot, M.~Geist, and O.~Pietquin, ``A cascaded supervised learning
  approach to inverse reinforcement learning,'' in \emph{Joint European
  conference on machine learning and knowledge discovery in databases}.\hskip
  1em plus 0.5em minus 0.4em\relax Springer, 2013, pp. 1--16.

\bibitem{blei2017variational}
D.~M. Blei, A.~Kucukelbir, and J.~D. McAuliffe, ``Variational inference: A
  review for statisticians,'' \emph{Journal of the American Statistical
  Association}, vol. 112, no. 518, pp. 859--877, 2017.

\bibitem{rubner2000earth}
Y.~Rubner, C.~Tomasi, and L.~J. Guibas, ``The earth mover's distance as a
  metric for image retrieval,'' \emph{International journal of computer
  vision}, vol.~40, no.~2, pp. 99--121, 2000.

\bibitem{frogner2015learning}
C.~Frogner, C.~Zhang, H.~Mobahi, M.~Araya, and T.~A. Poggio, ``Learning with a
  wasserstein loss,'' in \emph{Advances in Neural Information Processing
  Systems}, 2015, pp. 2053--2061.

\bibitem{villani2008optimal}
C.~Villani, \emph{Optimal transport: old and new}.\hskip 1em plus 0.5em minus
  0.4em\relax Springer Science \& Business Media, 2008, vol. 338.

\bibitem{he2018wasserstein}
R.~He, X.~Wu, Z.~Sun, and T.~Tan, ``Wasserstein cnn: Learning invariant
  features for nir-vis face recognition,'' \emph{IEEE transactions on pattern
  analysis and machine intelligence}, vol.~41, no.~7, pp. 1761--1773, 2018.

\bibitem{arjovsky2017wasserstein}
M.~Arjovsky, S.~Chintala, and L.~Bottou, ``Wasserstein gan,'' \emph{arXiv
  preprint arXiv:1701.07875}, 2017.

\bibitem{sutton1998reinforcement}
R.~S. Sutton and A.~G. Barto, \emph{Reinforcement learning: An
  introduction}.\hskip 1em plus 0.5em minus 0.4em\relax MIT press Cambridge,
  1998, vol.~1, no.~1.

\bibitem{alger2016irl}
\BIBentryALTinterwordspacing
M.~Alger, ``Inverse reinforcement learning,'' 2016. [Online]. Available:
  \url{https://doi.org/10.5281/zenodo.555999}
\BIBentrySTDinterwordspacing

\bibitem{brockman2016openai}
G.~Brockman, V.~Cheung, L.~Pettersson, J.~Schneider, J.~Schulman, J.~Tang, and
  W.~Zaremba, ``Openai gym,'' \emph{arXiv preprint arXiv:1606.01540}, 2016.

\bibitem{mnih2015human}
V.~Mnih, K.~Kavukcuoglu, D.~Silver, A.~A. Rusu, J.~Veness, M.~G. Bellemare,
  A.~Graves, M.~Riedmiller, A.~K. Fidjeland, G.~Ostrovski, \emph{et~al.},
  ``Human-level control through deep reinforcement learning,'' \emph{Nature},
  vol. 518, no. 7540, p. 529, 2015.

\end{thebibliography}
\bibliographystyle{IEEEtran} 

\end{document}